\documentclass[letterpaper, 10 pt, journal, twoside]{IEEEtran}

\ifCLASSINFOpdf
\else
\fi
\usepackage{graphics} 
\usepackage{epsfig} 
\usepackage{times} 
\usepackage{amsmath} 
\usepackage{amssymb}  
\usepackage{graphicx}
\usepackage{booktabs}
\usepackage{multirow}
\usepackage{mwe}
\usepackage{siunitx}
\usepackage{colortbl}
\usepackage{makecell}
\usepackage{float}
\usepackage{soul}

\usepackage[normalem]{ulem}
\usepackage{xcolor}
\definecolor{Blue}{rgb}{0,0,1}
\definecolor{Orange}{rgb}{0.929,0.49,0.192}
\definecolor{oursgray}{gray}{0.93}
\newcommand{\rd}[1]{#1}

\usepackage{placeins}
\usepackage{pifont}
\usepackage{hyperref}
\hypersetup{
    citecolor=green,
    colorlinks=true,
    linkcolor=blue,
    filecolor=magenta,      
    urlcolor=cyan,
    pdftitle={Overleaf Example},
    pdfpagemode=FullScreen,
    }
\DeclareMathOperator*{\TopKmin}{TopKmin}

\title{\LARGE \bf
Real-to-Sim for Highly Cluttered Environments via Physics-Consistent Inter-Object Reasoning
}

\author{Tianyi Xiang$^{1}$, Jiahang Cao$^{2}$, Sikai Guo$^{1}$, Guoyang Zhao$^{1}$, Andrew F. Luo$^{2}$, and Jun Ma$^{1}$
\thanks{Manuscript received: February 12, 2026; Revised April 21, 2026; Accepted May 13, 2026.}
\thanks{This paper was recommended for publication by Editor Aniket Bera upon evaluation of the Associate Editor and Reviewers' comments.
}
\thanks{$^{1}$Tianyi Xiang, Sikai Guo, Guoyang Zhao, and Jun Ma are with the Robotics and Autonomous Systems Thrust, The Hong Kong University of Science and Technology (Guangzhou), Guangzhou 511453, China (e-mail: txiang031@connect.hkust-gz.edu.cn; sguo837@connect.hkust-gz.edu.cn; gzhao492@connect.hkust-gz.edu.cn; jun.ma@ust.hk).}
\thanks{$^{2}$Jiahang Cao and Andrew F. Luo are with the Institute of Data Science, The University of Hong Kong, Hong Kong SAR, China (e-mail: jiahang@connect.hku.hk; aluo@hku.hk).}
\thanks{Digital Object Identifier (DOI): see top of this page.}
}

\begin{document}

\maketitle

\begin{abstract}

Reconstructing physically valid 3D scenes from single-view observations is a prerequisite for bridging the gap between visual perception and robotic control. However, in scenarios requiring precise contact reasoning, such as robotic manipulation in highly cluttered environments, geometric fidelity alone is insufficient. Standard perception pipelines often neglect physical constraints, resulting in invalid states, e.g., floating objects or severe inter-penetration, rendering downstream simulation unreliable. To address these limitations, we propose a novel physics-constrained Real-to-Sim pipeline that reconstructs physically consistent 3D scenes from single-view RGB-D data. Central to our approach is a differentiable optimization pipeline that explicitly models spatial dependencies via a contact graph, jointly refining object poses and physical properties through differentiable rigid-body simulation. Extensive evaluations in both simulation and real-world settings demonstrate that our reconstructed scenes achieve high physical fidelity and faithfully replicate real-world contact dynamics, enabling stable and reliable contact-rich manipulation. Our project page is at: \url{https://physics-constrained-real2sim.github.io/}.

\end{abstract}



\vspace{-6pt}
\section{INTRODUCTION}

\IEEEPARstart{C}{luttered} environments commonly arise in warehouses and factories, forming the basis of many robotic manipulation tasks, including object retrieval, pushing, and rearrangement \cite{correll2016analysis, dogar2011framework}.
In these settings, objects are coupled through complex contact and support relationships, making manipulation actions highly sensitive to inter-object interactions \cite{dogar2011framework, kartmann2018extraction}.
A central challenge is that robots must reason beyond the independent geometry of individual objects to infer inter-object physical dependencies, including support hierarchies, collision constraints, and frictional contacts \cite{kartmann2018extraction, xia2025holoscene}.
Real-to-Simulation (Real2Sim) \cite{villasevil2024reconciling} aims to bridge the gap between perception and control by replicating real-world dynamics in simulation, enabling reliable planning and policy learning through real-sim-real transfer \cite{chebotar2019closing}.
However, accurately modeling dynamically consistent inter-object relationships remains difficult, as it requires jointly reasoning about tightly coupled geometric and physical constraints \cite{kartmann2018extraction}, such as contact, friction, and non-penetration, that are highly nonlinear and only partially observable from a single visual observation.

\begin{figure}
    \centering
    \includegraphics[width=\linewidth]{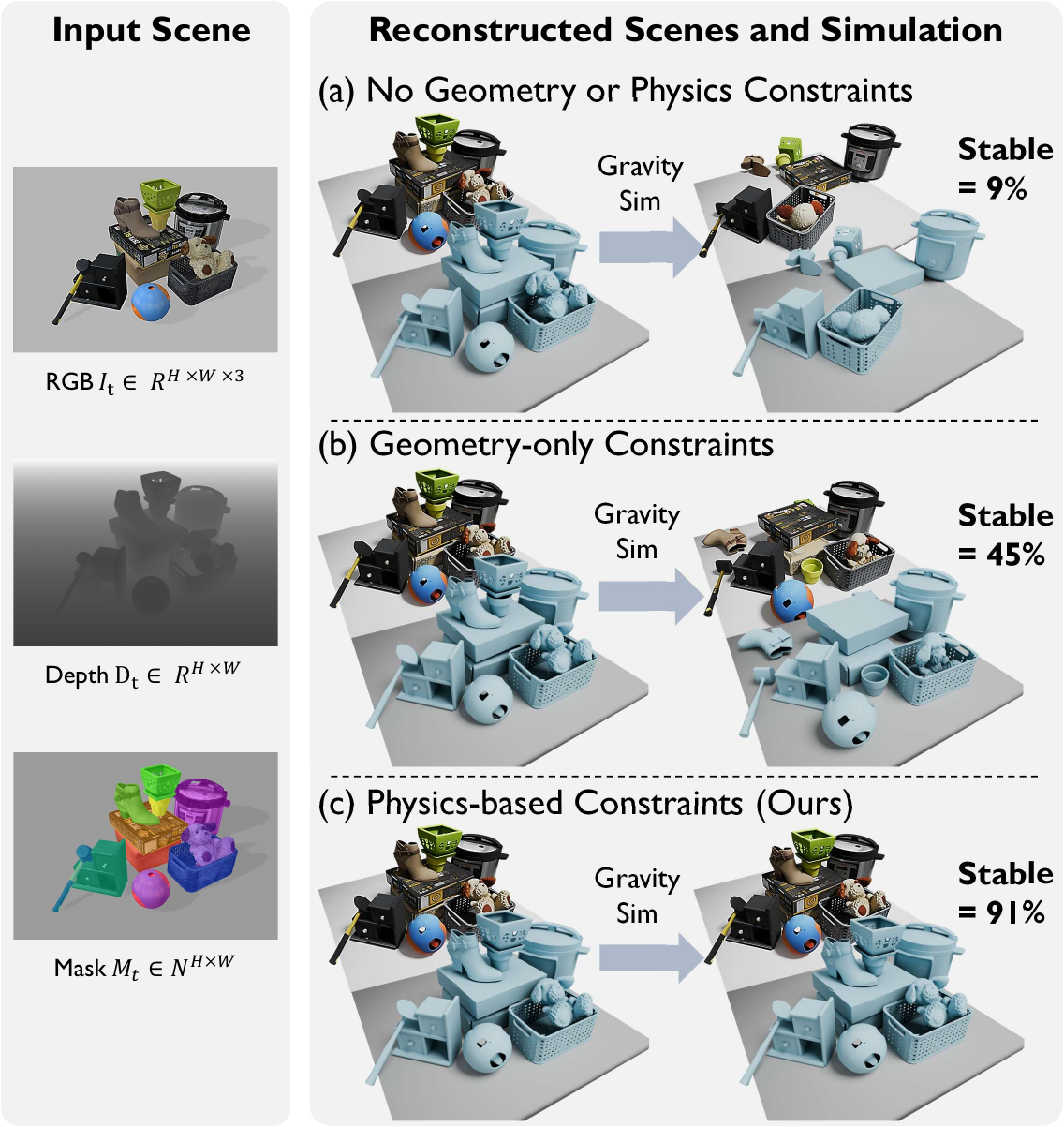}
    \vspace{-20pt}
    \caption{\textbf{Scene-level Real2Sim methods for physical stability.} Given a single RGB-D observation and instance masks, we reconstruct the scene and simulate in PyBullet \cite{coumans2016pybullet}. (a) SAM3D \cite{chen2025sam} with Iterative Closest Point (ICP) refinement, without geometric and physical constraints, results in interpenetration and floating, leading to unstable rollouts. (b) The geometry-only constrained method avoids penetration and ensures minimum contact, but does not guarantee long-horizon stability. (c) Our physics-constrained method jointly optimizes pose and physical parameters (e.g., friction, mass, and center of mass) to ensure the resulting simulation remains physically stable over time.}
    \vspace{-16pt}
    \label{fig1:spotLight}
\end{figure}

To incorporate physical reasoning into object reconstruction, prior works introduce physics-based knowledge into the regression pipeline \cite{ni2024phyrecon, guo2024physically}. These methods primarily focus on stabilizing individual objects with respect to the ground, and thus do not model the hierarchical spatial dependencies in highly cluttered multi-object scenes. 
Other approaches explicitly reason about inter-object relationships at the scene level. Geometry-based pose-refinement approaches such as CAST \cite{yao2025cast} and other related methods \cite{malenicky2025physpose, han2021reconstructing} enforce non-penetration and minimal contact using simple geometric heuristics that approximate physical constraints. While effective at resolving local interpenetration, these heuristics do not guarantee long-horizon dynamic stability under contact-rich interactions. 
Alternatively, sampling-based methods such as HoloScene \cite{xia2025holoscene} and related work \cite{song2018inferring} search for stable configurations using Monte Carlo Tree Search. Although effective, these approaches are computationally expensive, sample-inefficient, and generally require multi-view observations. 

From a complementary perspective, differentiable physics simulators provide an efficient and deterministic mechanism for inferring dynamic properties (e.g., initial states, contact parameters, and physical attributes) by minimizing discrepancies between simulated and observed trajectories \cite{freeman2021brax, strecke2021diffsdfsim, zhu2025one}. While this line of work does not explicitly target scene-level reconstruction, it offers a plausible way to encode physical consistency through gradient-based optimization. However, existing differentiable physics systems predominantly operate in single-object or single-agent settings, and extending them to multi-object scenes remains challenging due to highly coupled and nonlinear contact dynamics.

To bridge this gap, we propose a fully physics-constrained Real2Sim approach for generating dynamically consistent digital scenes by jointly optimizing both geometric and dynamic properties of highly cluttered objects from a single RGB-D image.
Our approach explicitly models inter-object spatial dependencies via a contact graph and organizes coupled contact constraints.
Physical plausibility is enforced by directly minimizing stability violations through differentiable physics, rather than relying on static geometric proxies as in prior works.
As shown in Fig. \ref{fig1:spotLight}, our reconstructed scenes maintain dynamic consistency over time compared with prior paradigms.
Our main contributions are as follows:

\begin{itemize}

    \item A Real2Sim framework that reconstructs a dynamically consistent digital twin of highly cluttered environments from a single RGB-D observation.

    \item A contact graph that explicitly captures inter-object support and contact relationships, and leverages these dependencies to guide hierarchical optimization.

    \item A hierarchical physics-constrained optimization strategy based on differentiable rigid-body simulation, which effectively resolves the highly coupled contact dynamics.

    \item Comprehensive validation through simulations and real-world experiments on a robotic platform, showcasing that the reconstructed scenes remain physically consistent and enable reliable simulation of contact-rich interactions.

\end{itemize}

\begin{figure*}
    \centering
    \includegraphics[width=0.95\textwidth]{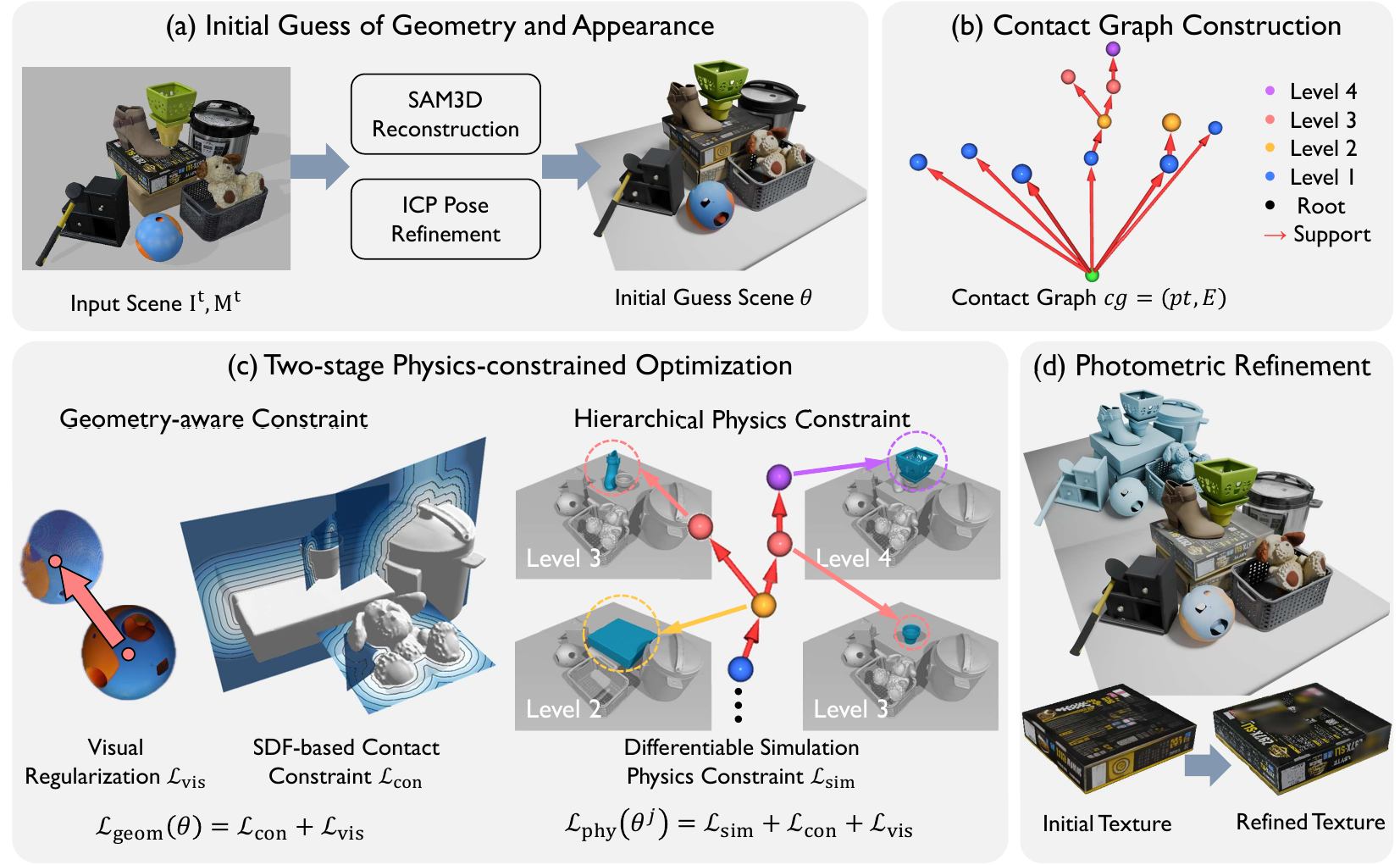}
    \vspace{-4pt}
    \caption{\textbf{Overview of our method.} Our physics-constrained Real2Sim pipeline consists of four stages. (a) \textbf{Initial Reconstruction:} Given a single RGB-D image $I_t$ and instance masks $M_t$, we obtain an initial estimation of objects geometry and appearance $\theta$ using SAM3D \cite{chen2025sam} and ICP pose refinement. (b) \textbf{Contact Graph Construction:} We construct a contact graph $cg = (pt, E)$, where parse tree $pt$ represents supporting tree and edges $E$ encode proximal relationships between objects. (c) \textbf{Two-Stage Physics-Constrained Optimization:} Guided by the contact graph, we optimize object properties in two stages. First, a geometry-aware optimization introduces SDF-based contact constraints and visual regularization to globally refine object poses, producing a penetration-free and contact-consistent initialization. Second, a hierarchical physics-constrained optimization, guided by the sequence of parse tree, uses differentiable simulation to jointly refine initial pose and physical parameters of each object for long-horizon physical stability. (d) \textbf{Photometric Refinement:} As a final post-process, object textures are refined using a differentiable renderer to achieve photometric consistency.}
    \vspace{-12pt}
    \label{fig:method}
\end{figure*}

\vspace{-6pt}
\section{RELATED WORK}

\subsection{Real2Sim Scene Reconstruction for Robotics}

Real2Sim scene reconstruction aims to convert limited observations into digital scenes that are both visually plausible and physically consistent.
Early approaches rely on manual scene design \cite{villasevil2024reconciling}, which can produce high-quality assets but are labor-intensive and lack scalability.
To improve scalability, asset-retrieval-based approaches \cite{dai2024acdc} match observations to pre-existing object models, but often sacrifice fidelity and practicality.
Moving beyond asset reuse, inverse-graphics methods infer 3D geometry from 2D observations by inverting the rendering process and leveraging shape priors \cite{chen2024urdformer}. 
More recently, neural scene representations, including Gaussian-splatting-based \textit{embodied Gaussians} \cite{abou2024physically, abou2025real} and its extensions with soft-material modeling \cite{jiang2025phystwin}, further improve scene visual fidelity. 
In parallel, foundation-model-driven approaches further enhance single-view completeness and visual plausibility \cite{huang2026synctwin, nasiriany2024robocasa}. 
However, despite their differences in representation and supervision, most existing methods are predominantly appearance-driven and do not enforce physical reasoning, resulting in dynamically unstable reconstructions that limit their applicability to contact-rich manipulation tasks.

\vspace{-6pt}
\subsection{Physics-Aware Reconstruction}

To enhance physical stability during reconstruction, prior works incorporate physics reasoning into the reconstruction pipeline. These methods can be broadly grouped according to the types of interactions they model: object–ground interactions and inter-object interactions. Object–ground interaction methods primarily optimize the stability of individual objects resting on the ground, often by incorporating differentiable physics engines as components within a regression pipeline \cite{ni2024phyrecon, guo2024physically}.
While effective at eliminating obvious violations such as floating or unsupported relations, these approaches often overlook inter-object dependencies and cannot model rigid-body coupled interactions.
On the other hand, inter-object interaction methods elevate reconstruction to scene-level physical reasoning.
Geometry-based pose refinement approaches, including CAST \cite{yao2025cast} and related works \cite{malenicky2025physpose, han2021reconstructing}, resolve interpenetration and minimal contact, but do not guarantee long-horizon dynamic stability. 
Sampling-based approaches, such as HoloScene \cite{xia2025holoscene} and prior work \cite{song2018inferring}, reason about stability via sampling-based optimization, at the cost of high computational complexity and sample inefficiency. 
These limitations arise because purely geometric constraints do not guarantee dynamic stability, while sampling-based strategies incur substantial computational overhead.
In contrast to these approaches, our work enforces physical consistency through deterministic, differentiable physics-based simulation, enabling dynamic stability to be optimized directly.

\vspace{-6pt}
\subsection{Differentiable Simulation}

Differentiable simulators enable end-to-end optimization by providing analytical gradients through rigid-body dynamics, contact, and friction, and have been widely used in robotics for system identification, model-based policy learning, and trajectory optimization \cite{freeman2021brax, strecke2021diffsdfsim, zhu2025one}. 
However, existing frameworks predominantly focus on single-object or single-agent scenarios, limiting their applicability to cluttered scenes with complex multi-object interactions. 
Scaling differentiable physics to enable scene-level reasoning remains challenging due to the rapidly increasing nonlinearity of contact dynamics as the number of interacting objects grows.
In contrast, we propose a hierarchical physics-constrained optimization framework that integrates differentiable simulation with an explicit contact graph, enabling scalable scene-level Real2Sim.

\vspace{-4pt}
\section{PROBLEM DEFINITION}

We first define our problem of Real2Sim for highly cluttered environments from a single RGB-D observation. At timestep $t$, the robot observes an RGB-D image $I_t \in  \mathbb{R}^{H \times W \times 4}$ and instance masks $M_t\in \mathbb{N}^{H \times W}$ (inferred or ground truth) from a calibrated camera with known intrinsics and extrinsics. We assume a static tabletop scene composed of multiple rigid objects placed on a support surface, where objects exhibit hierarchical inter-object relationships, including stacking, supporting, and leaning contacts.
Each object is parameterized by a set of physical and visual properties $\theta^i = \{q^i, c^i, m^i, f_c^i, T^i \}$, where $q^i \in SE(3)$, $c^i \in \mathbb{R}^3$, $m^i \in \mathbb{R}_{+}$, $f_c^i \in \mathbb{R}_{+}$, and $T^i$ denote the object pose, center of mass (COM), mass, surface friction coefficient, and texture, respectively. Our goal is to estimate the complete set of scene parameters $\theta^i$, thereby reconstructing a dynamically consistent and photorealistic digital twin scene. The scene is required to remain long-horizon statically stable under gravity when simulated with rigid-body dynamics.

As shown in Fig. \ref{fig:method}, our method proceeds in four stages. First, we extract the initial guess of geometry and appearance via SAM3D-objects \cite{chen2025sam} and ICP pose refinement. Next, we construct an explicit contact graph $ cg = (pt, E)$ to represent the inter-object relation, where $pt$ represents hierarchical supporting relations and $E$ encodes the proximal relation. Using the identified contact graph, we perform a two-stage physics-constrained optimization. The first stage conducts geometry-aware optimization to globally refine object poses $q$, producing a penetration-free and contact-consistent initialization. 
In the second stage, we build a hierarchical optimization graph guided by the parse tree $pt$. 
For each node, we perform physics-constrained optimization to object $\theta^i$ using a differentiable rigid-body simulator. The simulator function $g$ is defined as $q^{i+1} = g(q^{i}, f_{ext}^{i}, \theta^i) $, where $f_{ext}^i$ is the external force (e.g., gravity).
The differentiable optimization leverages the static scene as a zero-velocity prior, directly constraining all objects' simulated velocity $\dot q$ to be zero during long-horizon rollouts under gravity. Finally, the object texture $T^i$ is refined through a differentiable renderer for photometric consistency.

\vspace{-6pt}
\section{METHODOLOGY}

\subsection{Initial Guess of Geometry and Appearance}

Our method relies on a reasonable initial guess. Since the scene contains complex inter-object relations, the initialization method must be robust to obtain a reasonable estimation of both geometry and appearance from partial observations with heavy occlusion. We take advantage of a foundation reconstruction model, SAM3D-objects \cite{chen2025sam}, which can predict geometry, texture, and layout from a single RGB-D image. However, the mesh geometry and transforms predicted from SAM3D often deviate from the real world. To enhance the initial transforms, we perform ICP pose refinement implemented in Open3D \cite{Zhou2018}. Although penetration and floating cannot be fully resolved, the resulting geometry–appearance prior provides (1) a realistic approximation of complete mesh geometry, (2) a physically meaningful initialization for object poses, and (3) consistent texture estimates. 

\vspace{-6pt}
\subsection{Contact Graph Construction}

Given the geometry prior of the object set, we next infer their inter-object relations by constructing a contact graph $cg = (pt, E)$. As shown in Fig. \ref{fig:method}(b), the scene graph $cg$ can be encoded as a directed parse tree $pt=(V,S)$ representing vertical support structure, and an undirected edge set $E$ capturing proximal relations. 
Our contact graph formulation is inspired by the scene graph representation proposed in \cite{han2021reconstructing}, and is augmented with additional support-related attributes based on the support polygon criterion \cite{mcghee1968stability}, which serves as a necessary but not sufficient condition for static stability.

\textbf{Scene Entity Nodes.} The node set $V$ contains a ground node as the global root and a set of object nodes $v_i = \langle o_i, M_i, B_i, \Pi_i, F_p\rangle$. Each node includes the object instance label $o_i$, its triangular mesh $M_i$, an oriented bounding box (OBB) $B_i$, a set of upward-facing supporting planes $\Pi_i = { \pi_i^k }_{k=1}^{|\Pi_i|}$ obtained by selecting mesh faces whose normals are opposite to the gravity, and a support polygon $F_p$ defined as the convex hull of the mesh projected onto the $XY$ plane.

\textbf{Supporting Relations.} To determine the parent (supporting object) of each non-root node $v_c$, we perform a two-stage procedure consisting of broad-phase filtering followed by narrow-phase geometric likelihood estimation: (1) Broad-phase candidate selection: For each non-root object $v_c$, a feasible parent $v_p$ must lie vertically below the $v_c$, and their projected OBBs must overlap in the $XY$ plane; (2) Narrow-phase support likelihood: Given the possible candidate parents $v_p$, we compute a support likelihood combining the support gap and a simplified support polygon consistency:
\begin{equation}
H(v_c, v_p, \pi_p^k)= (1 - d_{\text{gap}}(v_c, \pi_p^k))\,(1 - d_{\text{com}}(\mathbf{{c}_c},F_p)),
\vspace{-2pt}
\end{equation}
where $d_{\text{gap}}(v_c, \pi_p^k)$ denotes the normalized support gap, defined as the minimum distance between the bottom surface of $v_c$ and the supporting plane $\pi_p^k$. Smaller gaps indicate more plausible support. The second term $d_{\text{com}}(\mathbf{{c}_c},F_p)$ measures the consistency between the projected COM $\mathbf{c}_c$ of $v_c$ and the support polygon $F_p$ of $v_p$. We compute $d_{\text{com}}(\mathbf{{c}_c},F_p)=\min_{\mathbf{x}\in F_p}\|\mathbf{{c}_c}-\mathbf{x}\|_2$. This distance is zero if the COM lies within the parent’s footprint (indicating stable support), and positive otherwise, reflecting the simplified standing consistency. The parent $v_p$ and plane $\pi_p^k$ with the highest likelihood define the directed support edge $s_{p,c} = \langle v_p, v_c\rangle$. 

\textbf{Proximity Edges.}
We add undirected edges $E$ between pairs of objects whose Axis-Aligned Bounding Boxes overlap. These edges represent all potential contact interactions, while narrow-phase is evaluated by SDF-based non-interpenetration constraints, as described in the following section.
\vspace{-4pt}
\subsection{Two-Stage Physics-Constrained Optimization}
\vspace{-2pt}

The initial reconstruction of a scene is often not physically plausible for downstream simulation. Such discrepancies arise not only from interpenetration or floating artifacts but also from inaccurate physical properties (e.g., mass, friction, and COM). To address these issues, we propose a two-stage physics-constrained optimization. In the first stage, we apply the geometry-aware constraint to globally regularize object poses, producing a penetration-free configuration with coherent contact support. In the second stage, we employ the hierarchical physics constraint to optimize object-level initial pose and physical parameters. \rd{Importantly, our hierarchical design should be viewed as an optimization-oriented approximation rather than relying on perfectly specified contact relationships. The selected parent mainly defines a reasonable parse tree and optimization order, while local ambiguity (e.g., multi-parent support or coupled leaning contacts) is further resolved through differentiable physics simulation.}

\subsubsection{Geometry-Aware Constraint}

We first globally optimize all object poses $q$ to resolve geometric inconsistency by minimizing an objective function that enforces contact constraints and visual alignment: 
\begin{equation}
\mathcal{L}_{\mathrm{geom}}(q)
=
\lambda_{c}\sum_{(i,j)\in cg} \mathcal{L}_{\mathrm{con}}(q^i, q^j)
+
\lambda_{v}\sum_i \mathcal{L}_{\mathrm{vis}}(I_t, q^i),
\end{equation}
where $\mathcal{L}_{\mathrm{con}}$ is the SDF-based contact constraint applied to object pairs specified by the contact graph $cg$. We define $\mathcal{L}_{\mathrm{con}}$ separately for supporting and proximal pairs. In addition, $\mathcal{L}_{\mathrm{vis}}$ is the visual regularization that anchors object poses to the initial estimates.

\textbf{Supporting Pair Constraints.} 
For an edge $(i,j)$ in the parse tree $pt$, we enforce non-penetration and sufficient multi-point contact such that the child object $j$ can stably rest on the parent object $i$. Let $D^i(p)$ denote the Signed Distance Function (SDF) of object $i$ evaluated at point $p$, and let $\mathcal{S}(q_j)$ denote the surface points sampled from object $j$ under pose $q_j$. The supporting pair loss is defined as:
\begin{equation}
\begin{aligned}
\mathcal{L}_{\mathrm{con}}(q^i,q^j) \; & =\;
\lambda_{\mathrm{pen}}
\left(
\frac{1}{N}\sum_{n=1}^{N}
\max_{x \in \mathcal{S}(q_j)}\!\big(0,\; -D^{i}(\mathbf{x})\big)
\right)
\; \\& +\;
\lambda_{\mathrm{con}}
\;\max\!\bigg(
0,\;
\frac{1}{K} \TopKmin_{x \in \mathcal{S}(q_j)} D^i(\mathbf{x}) 
\bigg),
\label{Equ:supporting}
\end{aligned}
\end{equation}
where the first term penalizes any penetration between two objects' surfaces. The second term encourages multi-point contact by ensuring that at least $K$ of the smallest SDF samples lie close to the contact surface. The point to mesh SDF is implemented by GPU accelerated Kaolin library \cite{KaolinLibrary}.

\textbf{Proximal Pair Constraints.} For object pairs $(i,j)$ connected only by proximity edges, we apply only the non-penetration term in (\ref{Equ:supporting}). This prevents collisions while allowing weak interaction, consistent with original physical roles.

\textbf{Visual Regularization.} This term anchors the optimization to the input observation by penalizing deviations of each object’s pose from its initial ICP-aligned estimate.


\subsubsection{Hierarchical Physics Constraint}

Although the scene refined by the geometry-aware constraints is generally not dynamically feasible due to simplified physical assumptions and unknown physical parameters, it resolves interpenetrating contacts and provides a suitable initialization for subsequent differentiable rigid-body simulation.

Due to the strong nonlinearity of rigid-body contact dynamics, joint optimization over all objects induces highly coupled dynamics, often leading to unstable gradients and poor convergence. To keep the problem computationally tractable, we optimize objects hierarchically, following the supporting tree $pt$ from the root node (e.g., the table) outward to leaf objects, as shown in Fig. \ref{fig:method}(c). This hierarchical formulation constitutes a sequential approximation, where parent objects are treated as locally fixed during child optimization. An object $j$ in the hierarchy is locally refined by solving the following optimization problem:
\begin{equation}
    \mathcal{L}_{phy}(\theta^j) = \lambda_{s}\mathcal{L}_{sim}(\theta^j) + \lambda_{c}\mathcal{L}_{con}(\theta^j) + \lambda_{v}\mathcal{L}_{vis}(I_t, \theta^j).
\end{equation}

To model the simulation stability loss $\mathcal{L}_{sim}$, we adopt DiffSDFSim \cite{strecke2021diffsdfsim}, a fully differentiable rigid-body simulator. DiffSDFSim is a velocity- and constraint-based time-stepping 3D simulation method that computes analytical gradients through contact dynamics formulated as a Linear Complementarity Problem (LCP). The dynamics function is defined as:
\begin{equation}
    (q_{i+1}, v_{i+1}) = g(q_i, v_i, \theta^j, f_{ext}, h_i),
\end{equation}
where $g$ denotes the differentiable dynamics operator that computes the next pose $q_{i+1}$ and velocity $v_{i+1}$ based on the current state $(q_i, v_i)$, object physical parameters $\theta^j$, external wrenches $f_{ext}$ (e.g., gravity), and the simulation time step $h_i$. Note that the velocity $v_i$ stacks both rotational and linear velocities. We then define $\mathcal{L}_{sim}$ as the time-integrated squared speed over the simulation horizon $T$:
\begin{equation}
\vspace{-1pt}
\min_{\theta^j}\quad \mathcal{L}_{sim}(\theta^j)
= \frac{1}{2}\sum_{i=0}^{T} \lVert v_i \rVert_2^2 \, h_i,
\vspace{-1pt}
\end{equation}
here, both the initial pose and physical parameters $\theta^j = \{q^j, c^j, m^j, f_c^j\}$ of object $j$ are jointly optimized. Minimizing this loss drives the object's velocity to zero, achieving a stable static equilibrium under gravity. Thanks to the differentiable formulation of frictional contact and collision constraints, DiffSDFSim \cite{strecke2021diffsdfsim} enables gradient backpropagation through complex contact interactions. This property allows efficient optimization of physical parameters, such as mass, friction coefficients, and the initial object pose, to maintain stable contact conditions during forward simulation.

To ensure the gradient backpropagation with respect to the COM, we correct the contact and friction Jacobians by expressing all contact point positions $p$ in the body-centric frame relative to the COM $c$, i.e., replacing $p$ with $p-c$. In addition, to enable efficient differentiable collision detection without computationally intractable convex decomposition, we adopt the Frank-Wolfe algorithm \cite{macklin2020local}, which operates directly between SDF and mesh. Accordingly, each optimized object $j$ is represented as a $128 \times 128 \times 128$ voxelized SDF using the GPU-accelerated Mesh2SDF in the Kaolin library \cite{KaolinLibrary}.

\vspace{-4pt}
\subsection{Photometric Refinement}

Due to the inherent ambiguity of single-view reconstruction, the initial texture estimates often lack high-frequency details. As shown in Fig. \ref{fig:method}(d), to improve the photometric fidelity, we optimize each object texture $T$ using a differentiable renderer DIB-R \cite{chen2019learning}, while keeping geometric parameters fixed. Since only a single view is provided, we incorporate an isotropic Total Variation (TV) regularizer to propagate photometric gradients across the texture map. Given the optimized geometry and pose, we render the object texture $T$ to obtain $I_r$ and compare it with the observed image $I_t$ over the region $\Omega \in M_j$. The objective is defined as:
\begin{equation}
\mathcal{L} = 
\frac{\lambda_p}{|\Omega|}
\sum_{p\in \Omega}
| I_r(p)-I_t(p) |
+
\frac{\lambda_t}{N}
\sum_{i,j}
\sqrt{
(\nabla_x T)^2 +
(\nabla_y T)^2 },
\end{equation}
where the first photometric term aligns the rendered image and observation, while the second TV term spreads gradients across the texture map. The summation is over all texture pixels $(i,j)$, and $N$ is the total number of texture pixels.

\begin{figure*}
    \centering
    \includegraphics[width=0.95\textwidth]{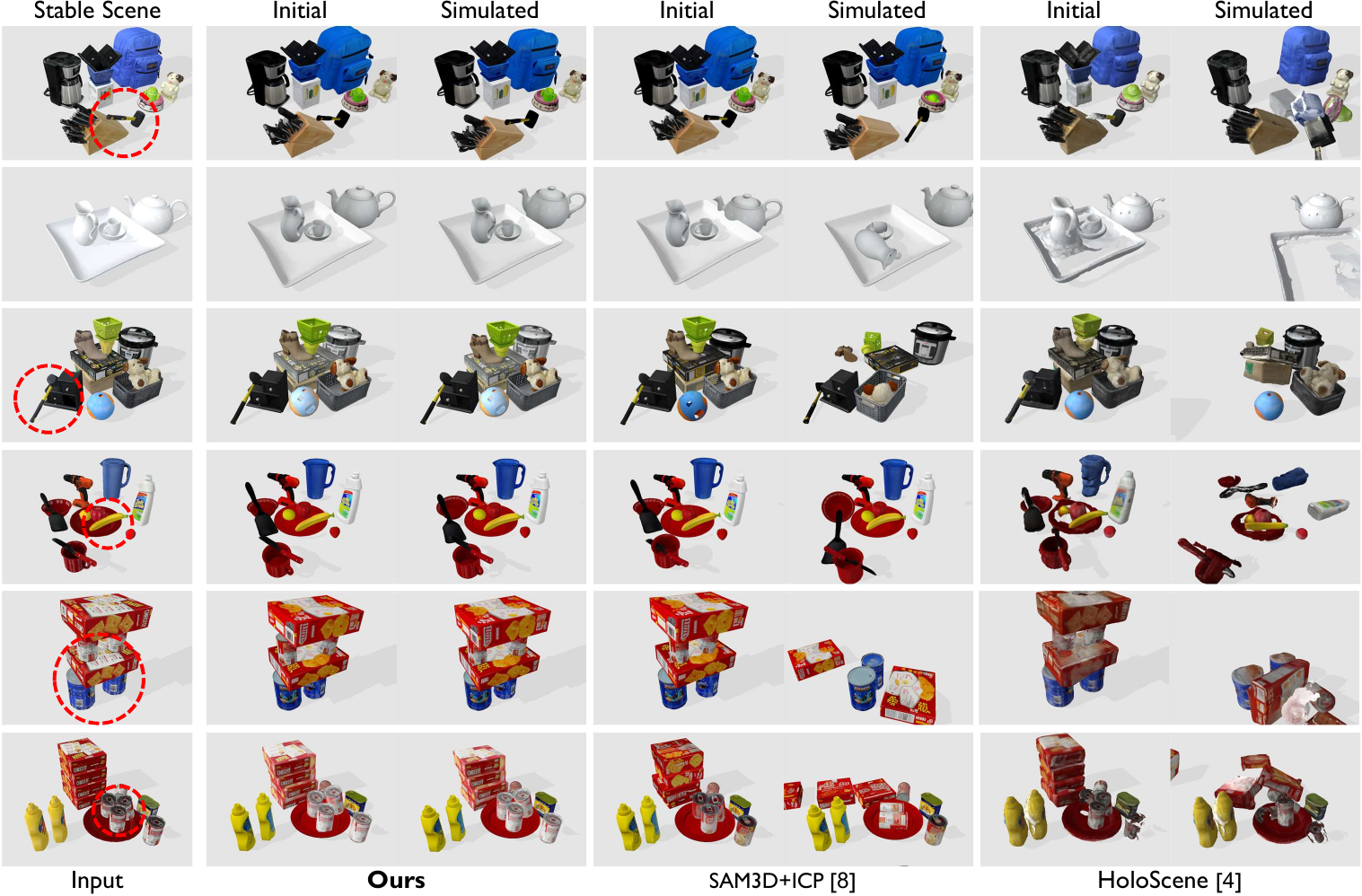}
    \vspace{-8pt}
    \caption{\textbf{Qualitative comparisons of physical simulation results with state-of-the-art scene-level reconstruction methods in the simulation environment.} We visualize geometry and appearance before and after physical simulation with gravity in PyBullet \cite{coumans2016pybullet}. \rd{Complex multi-contact structures, such as a board resting on two blocks and leaning against other objects, are highlighted with red dashed lines.}
    Our method produces non-interpenetrating, contact-coherent geometry and achieves long-horizon physical stability compared with baseline methods. 
    }
    \vspace{-14pt}
    \label{fig:sim_result}
\end{figure*}


\begin{table*}
  \centering
  \caption{Quantitative results of reconstruction on runtime, physical stability, geometry, and rendering metrics.}
  \vspace{-10pt}
  \resizebox{\linewidth}{!}{
  \begin{tabular}{
    llc
    S[table-format=4.1]   
    S[table-format=2.1]
    S[table-format=1.2]
    S[table-format=1.3]
    S[table-format=1.3]
    S[table-format=2.1]
    S[table-format=1.2]
    S[table-format=1.2]
    S[table-format=2.1]
    S[table-format=1.2]
    S[table-format=1.2]
  }
    \toprule
    &&& \multicolumn{1}{c}{\textbf{\rd{Time}}} 
       & \multicolumn{4}{c}{\textbf{Physical Stability}} 
       & \multicolumn{3}{c}{\textbf{Geometry}}
       & \multicolumn{3}{c}{\textbf{Rendering}} \\
    \cmidrule(lr){4-4}
    \cmidrule(lr){5-8}
    \cmidrule(lr){9-11}
    \cmidrule(lr){12-14}

    Dataset & Input & Method
      & {\rd{[min]}$\downarrow$}
      & {Stab. [\%]$\uparrow$}
      & {Set. [s]$\downarrow$}
      & {$V_t$ [m/s]$\downarrow$}
      & {$\Omega_t$ [rad/s]$\downarrow$}
      & {CD [mm]$\downarrow$} & {FS$\uparrow$} & {IoU$\uparrow$}
      & {PSNR$\uparrow$} & {SSIM$\uparrow$} & {LPIPS$\downarrow$} \\
    \midrule

    \multirow{4}{*}{GSO}
      & Image & \textbf{Ours}      
      & \rd{22.0}
      & \textbf{85.7} & \textbf{0.73} & \textbf{0.005} & \textbf{0.033} 
      & 30.8 & 0.56 & 0.69 
      & 20.8 & 0.84 & 0.15 \\
      & Image & {\scriptsize SAM3D+ICP \cite{chen2025sam}} 
      & \rd{10.0}
      & 51.4 & 1.48 & 0.078 & 0.338 
      & 19.7 & 0.79 & 0.82 
      & \textbf{22.5} & \textbf{0.87} & \textbf{0.11} \\
    \cmidrule(lr){2-14}
      & Video & HoloScene \cite{xia2025holoscene}
      & \rd{840.0}
      & 51.8 & 2.72 & 0.056 & 0.245 
      & \textbf{18.9} & \textbf{0.80} & \textbf{0.82} 
      & 18.1 & 0.79 & 0.18 \\
    \midrule

    \multirow{4}{*}{YCB}   
      & Image & \textbf{Ours}      
      & \rd{22.0}
      & \textbf{89.3} & \textbf{0.43} & \textbf{0.003} & \textbf{0.032} 
      & 43.0 & 0.66 & 0.68 
      & 21.1 & 0.86 & 0.12 \\
      & Image & {\scriptsize SAM3D+ICP \cite{chen2025sam}}
      & \rd{10.0}
      & 48.8 & 1.39 & 0.027 & 0.340 
      & 24.7 & 0.85 & 0.83 
      & \textbf{23.6} & \textbf{0.89} & \textbf{0.08} \\
    \cmidrule(lr){2-14}
      & Video & HoloScene \cite{xia2025holoscene}
      & \rd{840.0}
      & 49.2 & 2.67 & 0.021 & 0.201 
      & \textbf{14.8} & \textbf{0.90} & \textbf{0.88} 
      & 19.4 & 0.79 & 0.19 \\
    \bottomrule
  \end{tabular}
  }
\vspace{-14pt}
\label{tab:sim_result}
\end{table*}

\begin{table}
  \centering
  \small
  \caption{Ablation study of different constraint configurations in simulation experiments.}
  \vspace{-8pt}
  \setlength{\tabcolsep}{3pt} 
  \begin{tabular}{
    l
    cc
    S[table-format=2.1]
    S[table-format=1.2]
    S[table-format=1.3]
    S[table-format=1.2]
    S[table-format=2.1]
    S[table-format=1.2]
  }
    \toprule
    Dataset
      & Geo.
      & Phy
      & {Stab.$\uparrow$}
      & {Set.$\downarrow$}
      & {$V_t$$\downarrow$}
      & {$\Omega_t$$\downarrow$}
      & {CD$\downarrow$}
      & {FS$\uparrow$} \\
    \midrule
    \multirow{3}{*}{GSO}
      & \textcolor[HTML]{B22222}{\ding{55}} & \textcolor[HTML]{B22222}{\ding{55}} & 51.4 & 1.48 & 0.078 & 0.33 & \textbf{19.7} & \textbf{0.79} \\
      & \textcolor[HTML]{228B22}{\ding{51}} & \textcolor[HTML]{B22222}{\ding{55}} & 58.3 & 1.51 & 0.031 & 0.11 & 24.1 & 0.65 \\
      & \textcolor[HTML]{228B22}{\ding{51}} & \textcolor[HTML]{228B22}{\ding{51}} & \textbf{85.7} & \textbf{0.73} & \textbf{0.005} & \textbf{0.03} & 30.8 & 0.56\\
    \midrule
    \multirow{3}{*}{YCB}
      & \textcolor[HTML]{B22222}{\ding{55}} & \textcolor[HTML]{B22222}{\ding{55}} & 48.8 & 1.39 & 0.027 & 0.34 & \textbf{24.7} & \textbf{0.85} \\
      & \textcolor[HTML]{228B22}{\ding{51}} & \textcolor[HTML]{B22222}{\ding{55}} & 61.4 & 1.20 & 0.014 & 0.09 & 25.7 & 0.76 \\
      & \textcolor[HTML]{228B22}{\ding{51}} & \textcolor[HTML]{228B22}{\ding{51}} & \textbf{89.3} & \textbf{0.43} & \textbf{0.003} & \textbf{0.03} & 43.0 & 0.66 \\
    \bottomrule
  \end{tabular}
\vspace{-20pt}
\label{tab:ablation}
\end{table}

\begin{figure*}[t]
    \centering
    \includegraphics[width=0.9\textwidth]{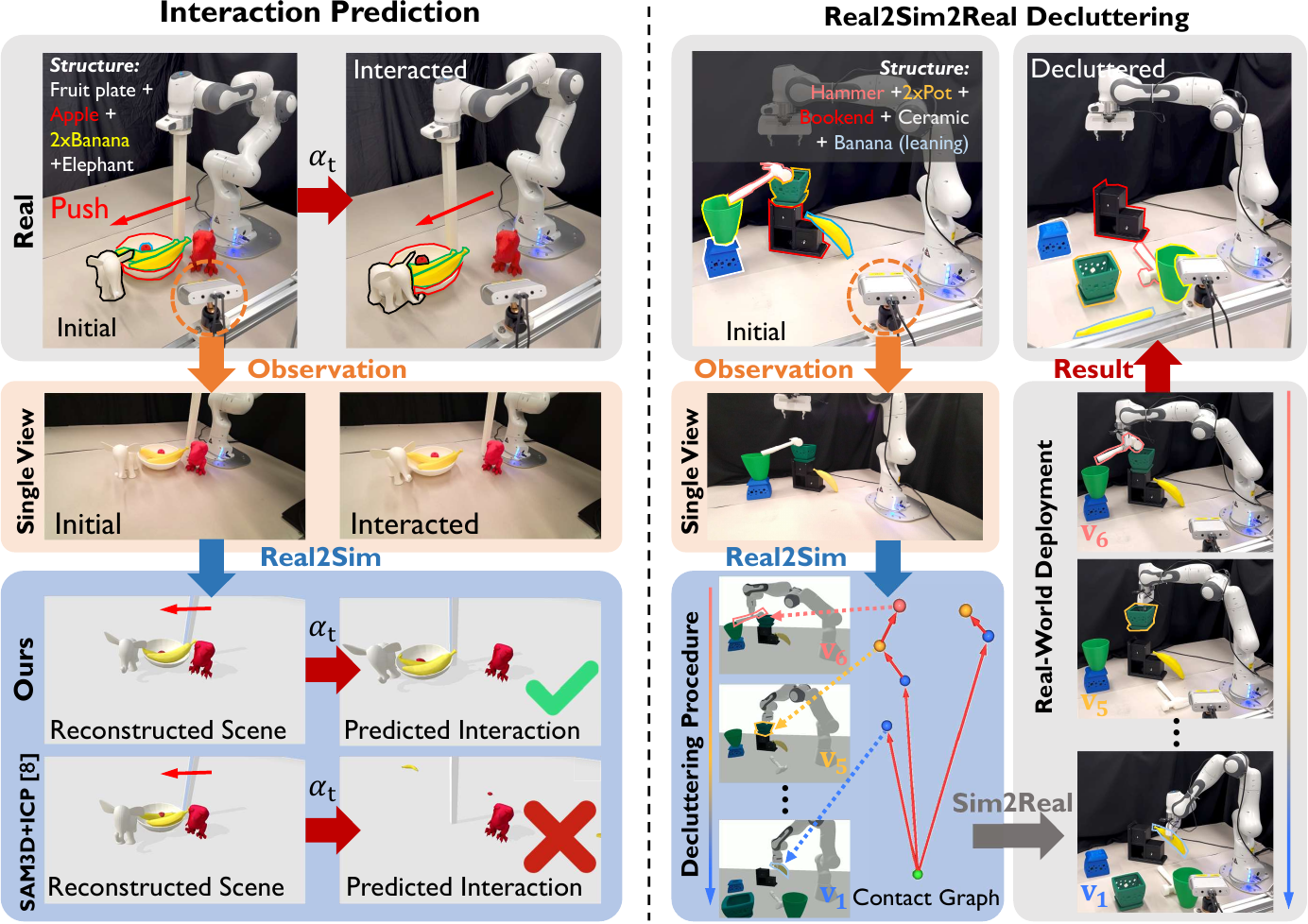}
    \vspace{-8pt}
    \caption{\rd{\textbf{Real-world experiments on interaction prediction and downstream robot decluttering.} Left: From a single RGB-D observation, our method reconstructs a physically consistent scene and predicts the outcome of a pushing interaction via simulation, achieving more accurate post-interaction states than SAM3D+ICP. Right: We further evaluate our method in a Real2Sim2Real decluttering, where the reconstructed scene with an inferred contact graph enables simulation-guided grasping sequence selection, and the strategy is successfully executed on the real robot to remove clutter.}}
    \vspace{-8pt}
    \label{fig:real_result}
\end{figure*}

\begin{table*}[t]
  \centering
  \small
  \setlength{\tabcolsep}{2pt}
  \caption{\rd{Quantitative results on physical stability, geometry, rendering metrics, interaction prediction, and Real2Sim2Real decluttering in the real-world experiment.}}
  \vspace{-2pt}
  \resizebox{\textwidth}{!}{%
  \begin{tabular}{
    l
    l
    S[table-format=2.1]
    S[table-format=1.2]
    S[table-format=1.3]
    S[table-format=1.2]
    S[table-format=2.1]
    S[table-format=2.1]
    S[table-format=1.2]
    S[table-format=1.2]
    S[table-format=2.1]
    S[table-format=2.1]
  }
    \toprule
    Dataset
      & Method
      & \multicolumn{4}{c}{Physical Stability}
      & \multicolumn{1}{c}{Geometry}
      & \multicolumn{3}{c}{Rendering}
      & \multicolumn{1}{c}{\rd{Interaction Prediction}}
      & \multicolumn{1}{c}{\rd{Decluttering}} \\
    \cmidrule(lr){3-6}
    \cmidrule(lr){7-7}
    \cmidrule(lr){8-10}
    \cmidrule(lr){11-11}
    \cmidrule(lr){12-12}
      &
      & {Stab. [\%] $\uparrow$}
      & {Set. [s] $\downarrow$}
      & {$V_t$ [m/s]$\downarrow$}
      & {$\Omega_t$ [rad/s]$\downarrow$}
      & {$\text{UCD}$ [mm]$\downarrow$}
      & {PSNR$\uparrow$}
      & {SSIM$\uparrow$}
      & {LPIPS$\downarrow$}
      & {\rd{$\text{UCD}$ [mm]$\downarrow$}}
      & {\rd{SR [\%] $\uparrow$}} \\
    \midrule

    \multirow{2}{*}{GSO}
      & \textbf{Ours}
      & \textbf{71.6}
      & \textbf{0.52}
      & \textbf{0.005}
      & \textbf{0.05}
      & 36.3
      & 19.5
      & 0.89
      & 0.11
      & \rd{\textbf{59.0}}
      & \textbf{\rd{70.0}} \\
      & {SAM3D+ICP \cite{chen2025sam}}
      & 00.0
      & 2.36
      & 0.141
      & 1.10
      & \textbf{20.0}
      & \textbf{21.5}
      & \textbf{0.92}
      & \textbf{0.08}
      & \rd{N/A}
      & \rd{00.0} \\

    \midrule

    \multirow{2}{*}{Toy4K}
      & \textbf{Ours}
      & \textbf{73.3}
      & \textbf{0.69}
      & \textbf{0.008}
      & \textbf{0.08}
      & 39.7
      & 17.8
      & 0.91
      & 0.09
      & \rd{\textbf{62.6}}
      & \rd{\textbf{80.0}} \\
      & {SAM3D+ICP \cite{chen2025sam}}
      & 15.0
      & 2.00
      & 0.053
      & 1.53
      & \textbf{21.3}
      & \textbf{20.93}
      & \textbf{0.94}
      & \textbf{0.06}
      & \rd{N/A}
      & \rd{00.0} \\

    \bottomrule
  \end{tabular}
  }

  \vspace{-15pt}
  \label{tab:real_world}
\end{table*}

\vspace{-8pt}
\section{Experiment}

\subsection{Simulation Experiments}

All simulation experiments are conducted in the PyBullet physics engine \cite{coumans2016pybullet}, as illustrated in Fig.~\ref{fig:sim_result}. We use object assets from the Google Scanned Objects (GSO) dataset \cite{downs2022google}, and the YCB object set \cite{calli2015benchmarking}. For each dataset, we construct 20 highly cluttered tabletop scenes. Each scene contains between 4 and 15 objects, randomly sampled from the dataset and manually arranged to form diverse inter-object relationships, including supporting, leaning, and close-proximity contacts.

To assess physical stability, reconstructed scenes are re-simulated in PyBullet under gravity. All physical properties are automatically estimated through our optimization process. \rd{We report the average computation time per scene as an efficiency measurement.} We report four stability metrics: stability ratio (Stab.), settling time (Set.), average translational velocity ($V_t$), and average rotational velocity ($\Omega_t$). The stability ratio is defined as: $\text{Stab.} \% = \frac{\text{stable instances}}{\text{all instances}}$, where an instance is considered stable if the translational pose change is below 5 cm and the Euler angles change is below $5^\circ$ over the rollout horizon. We also report the geometry quality with Chamfer Distance (CD), F-Score (FS), and Intersection over Union (IoU), and rendering quality using PSNR, SSIM, and LPIPS.

We compare our approach with representative instance-aware 3D scene reconstruction methods. For single-image input, we adopt SAM3D+ICP \cite{chen2025sam}, which combines a recent foundation model with ICP-based pose refinement. We further include a video-based baseline, HoloScene \cite{xia2025holoscene}, which performs sampling-based physical reasoning via Monte Carlo Tree Search. \rd{All experiments are conducted on a PC with an Intel i9-7900X CPU and a NVIDIA RTX 3080 GPU (10GB VRAM), whereas HoloScene \cite{xia2025holoscene} is evaluated on an RTX 4090 GPU (24GB VRAM). Our optimization time increases with the number of objects and the depth of the contact hierarchy. For scenes with 4–6 objects, the average runtime is 5 mins, while for scenes with 14–15 objects, it increases to 20 mins. Geometry-aware pose optimization takes about 10\% of the runtime, while physics-based joint optimization accounts for the remaining 90\%.}

As summarized in Table~\ref{tab:sim_result}, our method achieves the highest physical stability across both datasets while maintaining competitive geometry and rendering quality. Unlike SAM3D+ICP, which does not incorporate physical reasoning, our framework improves the stability ratio by an average of $38\%$, and consistently reduces both translational and rotational velocities to near zero, indicating convergence to a stable static equilibrium. On the other hand, HoloScene \cite{xia2025holoscene} removes all geometric inter-penetration after its hypothesis sampling stage, which often degrades local geometry near contact regions, especially in densely populated scenes where multiple objects interact (e.g., a fruit plate in the fourth row in Fig. \ref{fig:sim_result}).

\vspace{-5pt}
\subsection{Ablation Study}

Table \ref{tab:ablation} presents an ablation study analyzing the effect of geometric (Geo.) and physics (Phy.) constraints. Adding the geometric constraint yields a moderate improvement in physical stability and significantly reduces translational and rotational velocities. This indicates the geometry constraints successfully refine the clutter to be non-interpenetrating, but are generally not dynamically feasible. Building on this initialization, incorporating the hierarchical physics constraint leads to a substantial gain in stability and faster settling. \rd{We also observe a trade-off between physical stability and geometric accuracy. Under occluded single-view, the reconstructed meshes may be visually plausible while deviating substantially from the ground truth geometry. The point-wise refinement ICP aims to minimize registration error, but overfits these imperfect meshes and yields interpenetration or floating artifacts. In contrast, our physics-constrained optimization acts as a strong physical prior: it may move objects away from the geometry-only optimum to enforce dynamically feasible configurations. As a result, geometric metrics (e.g., CD) can slightly degrade, while physical stability improves substantially. Importantly, this trade-off is preferable for downstream data collection, since prioritizing geometric alignment alone can lead to unstable scenes like blowing up in the simulator, whereas geometry errors can be mitigated through domain randomization.}

\vspace{-8pt}
\subsection{Physical Experiments}

\rd{As shown in Fig. \ref{fig:real_result}, we first evaluate the real-world generalization of our method and then apply it in two robotics tasks: interaction outcome prediction and downstream object decluttering.}
\rd{In both settings, a Franka robotic arm and a static overhead Azure Kinect DK capture a single RGB-D observation of a cluttered tabletop scene.}
We randomly select and 3D print ten object assets from each of the GSO \cite{downs2022google} and Toy4K \cite{stojanov2021using} datasets. 
\rd{Each scene is manually assembled and contains 4-7 objects.} Due to the single-view observation setting, we only adopt SAM3D+ICP \cite{chen2025sam} as the baseline method. All optimization settings are identical to those used in simulation experiments. 
Since ground-truth object poses are unavailable and only partial depth measurements are observed, \rd{we quantify reconstruction quality and post-interaction prediction error using unilateral Chamfer Distance (UCD) between the observed object point cloud and the reconstructed meshes.}


We report the quantitative results of scene reconstruction in Table \ref{tab:real_world}. Our method achieves approximately $72\%$ stability and an average initial UCD of 40 mm. 
The high stability of the reconstructed initial scene enables \rd{interaction prediction} for pushing actions. As illustrated in \rd{Fig. \ref{fig:real_result}(left)}, our method successfully predicts scene evolution after replaying the real robot trajectory $a_t$. In contrast, SAM3D+ICP often fails due to severe object–object interpenetration at initialization, \rd{resulting in blowing up prior to end-effector interaction.} While the physical stability of our method drops by approximately $10\%$ compared to simulation experiments, the geometric reconstruction accuracy and rendering quality remain comparable. We attribute this gap primarily to noise and depth inaccuracies inherent in real-world depth cameras, which degrade ICP-based pose alignment. 
\rd{We further evaluate our method in a downstream Real2Sim2Real decluttering task. As shown in Fig. \ref{fig:real_result}(right), our method reconstructs a physically stable scene and infers a contact graph for decluttering sequence selection.
The scene reconstruction is performed only once at initialization. With the reconstructed complete object meshes, grasp poses are pre-annotated on the meshes, and simple IK-based trajectories using PyBullet \cite{coumans2016pybullet} are successfully executed in the real world.
This result provides stronger evidence that physically consistent scene reconstruction can serve as an effective interface between perception and robot action.}

\vspace{-6pt}
\section{Conclusion}

We propose a fully physics-constrained Real2Sim method that generates physically consistent digital twins for highly cluttered scenes from a single RGB-D observation.
Our method explicitly models inter-object support and contact relations via a hierarchical contact graph and refines object pose and physical properties (e.g., mass, friction, and COM) through differentiable physics-based optimization.
Experiments in both simulation and the real world show substantial gains in physical stability while maintaining competitive geometric and visual fidelity.
Real robot experiments further demonstrate that the reconstructed scenes enable reliable simulation of contact-rich pushing interactions.

\rd{\textbf{Limitations.}
Although our method can recover a physically plausible scene given a reasonable initialization, it remains limited by severe occlusions and single-view ambiguity. For example, a small object may be tightly sandwiched between two larger objects, leaving only a tiny visible region while its contact relationships remain unobserved. In such cases, the optimization may get trapped in a poor local minimum, yielding solutions that are physically stable yet visually implausible. In addition, our method does not consider advanced appearance models, complex lighting, or shadows under single-view supervision.}
\vspace{-6pt}

\bibliographystyle{IEEEtran}
\bibliography{explore_ideas}

\end{document}